\pdfoutput=1

\documentclass[11pt]{article}

\usepackage[dvipsnames]{xcolor}
\usepackage{authblk}
\usepackage{graphicx}
\usepackage{enumitem}
\usepackage{xspace}
\usepackage{xparse}
\DeclareDocumentCommand\newstep{o}{%
\item\IfNoValueTF{#1}{}{#1 \textendash\xspace}}
\newlist{steps}{enumerate}{1}
\setlist[steps]{label=\textit{RQ \arabic*:},leftmargin=*}

\usepackage[final]{acl}

\usepackage{times}
\usepackage{latexsym}

\usepackage{amssymb}
\usepackage{tablefootnote}
\usepackage{amsmath}
\usepackage{hyperref}
\usepackage{cleveref}
\crefrangelabelformat{section}{#3#1#4--#5\crefstripprefix{#1}{#2}#6}
\usepackage{float}

\usepackage[T1]{fontenc}

\usepackage[utf8]{inputenc}

\usepackage{booktabs}
\usepackage{multirow}
\usepackage{makecell} 

\usepackage{microtype}

\usepackage{inconsolata}

\newcommand{\wtq}{\textsc{WTQ}\xspace}
\newcommand{\wikisql}{\textsc{WikiSQL}\xspace}
\newcommand{\squall}{\textsc{Squall}\xspace}

\newcommand{\robut}{\textsc{RobuT}\xspace}

\newcommand{\tapex}{\textsc{TaPEx}\xspace}
\newcommand{\omnitab}{OmniTab\xspace}
\newcommand{\otableqa}{Table QA\xspace}
\newcommand{\tableqa}{E2E TQA\xspace}
\newcommand{\texttosql}{Text-to-SQL\xspace}
\newcommand{\syntableqa}{\textsc{SynTQA}\xspace}

\newcolumntype{L}[1]{>{\raggedright\let\newline\\\arraybackslash\hspace{0pt}}m{#1}}
\newcolumntype{C}[1]{>{\centering\let\newline\\\arraybackslash\hspace{0pt}}m{#1}}
\newcolumntype{R}[1]{>{\raggedleft\let\newline\\\arraybackslash\hspace{0pt}}m{#1}}

\title{\syntableqa: Synergistic Table-based Question Answering \\ via Mixture of Text-to-SQL and E2E TQA}

\author{Siyue Zhang$^{\diamondsuit \heartsuit}$ \quad Anh Tuan Luu$^{\diamondsuit}$ \quad Chen Zhao$^{\spadesuit \clubsuit}$}
\affil{
 $^{\diamondsuit}$Nanyang Technological University \quad $^{\spadesuit}$NYU Shanghai \\ $^{\heartsuit}$Alibaba-NTU Singapore Joint Research Institute \\
 $^{\clubsuit}$Center for Data Science, New York University
 \\\texttt{siyue001@e.ntu.edu.sg, anhtuan.luu@ntu.edu.sg, cz1285@nyu.edu}}

\begin{document}
\maketitle
\begin{abstract}
 Text-to-SQL parsing and end-to-end question answering (\tableqa) are two main approaches for Table-based Question Answering task. Despite success on multiple benchmarks, they have yet to be compared and their synergy remains unexplored. In this paper, we identify different strengths and weaknesses through evaluating state-of-the-art models on benchmark datasets: \texttosql demonstrates superiority in handling questions involving arithmetic operations and long tables; \tableqa excels in addressing ambiguous questions, non-standard table schema, and complex table contents. To combine both strengths, we propose a Synergistic Table-based Question Answering approach that integrate different models via answer selection, which is agnostic to any model types. Further experiments validate that ensembling models by either feature-based or LLM-based answer selector significantly improves the performance over individual models. Code will be publicly available at \url{https://github.com/siyue-zhang/SynTableQA}.
 
\end{abstract}

\section{Introduction}



\otableqa (\textsc{TQA}) takes a question and a table, and finds an answer based on the evidence from the table~\cite{pasupat-compositional}. With the help of large scale datasets~\cite{wikisql, yu-spider, yu-sparc, squall}, state-of-the-art (SOTA) \textsc{TQA} systems primarily focus on two approaches: \textit{semantic parsing} (\texttosql) that predicts a SQL query as intermediate semantic representation of the question, and then executes the SQL to find the answer \citep{wang-rat, scholak-picard, graphix}; \textit{end-to-end system} (\tableqa) that directly generates the answer from models pre-trained on table corpora, imitating human-like reasoning on questions and tables \citep{pasupat-compositional, iyyer-search, gupta-temptabqa}. Despite serving for a similar purpose, it's unclear what advantages these approaches have and their potential synergy.

\begin{figure}[t]
    \centering
    \includegraphics[width=0.5\textwidth]{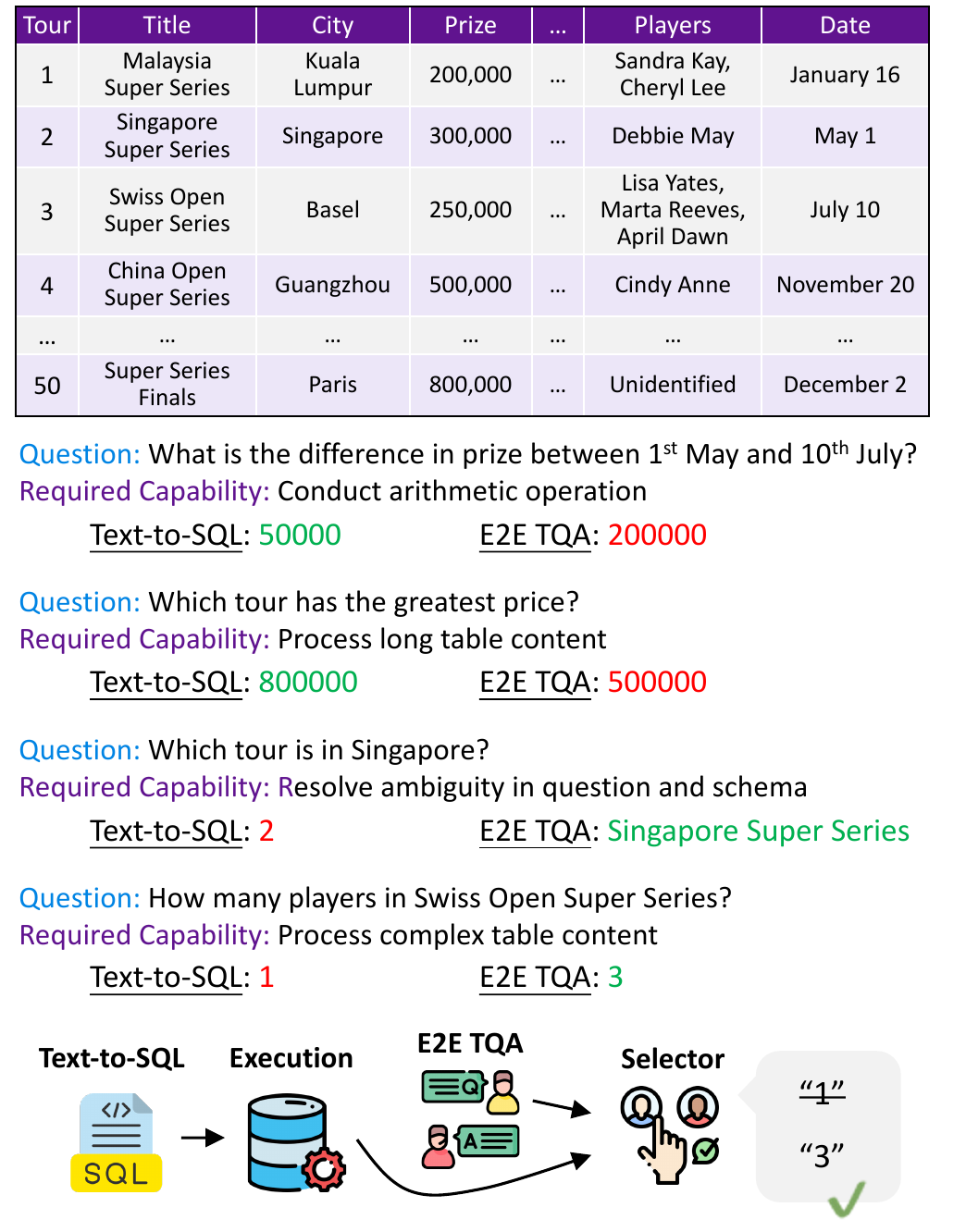}
    \caption{A demonstration of SOTA \otableqa models' strengths in solving different types of table-based questions, followed by an overview of \syntableqa. In a synergistic way, \syntableqa aggregates candidate answers from \texttosql and \tableqa models, and then select the final answer. The answers in green color are the correct answers.}
    \label{fig:flow}
\end{figure}


To answer these questions, we first (re)evaluate SOTA \texttosql models, i.e., T5 \cite{t5}, GPT \citep{OpenAI_GPT4_2023}, and DIN-SQL \citep{din}, as well as \tableqa models, i.e., \tapex \citep{tapex}, \omnitab \citep{jiang-omnitab}, and GPT, on benchmark datasets \wtq \citep{pasupat-compositional} and \wikisql \citep{wikisql}. The experiments show that while both \texttosql and \tableqa approaches are adept at simple questions, they have complementary strengths for complex questions and tables, as shown in Figure~\ref{fig:flow} (top): \texttosql is more proficient in numerical reasoning and processing long tables while \tableqa is better at ambiguous questions, complex schema and contents.

Motivated by their distinct strengths, 
we propose \textbf{Syn}ergistic \textbf{T}able-based \textbf{Q}uestion \textbf{A}nswering (\textbf{\syntableqa}, bottom of Figure \ref{fig:flow}), which aims to integrate the strengths of both models via answer selection. At each time, given the input of table, question, \texttosql answer, and \tableqa answer, the answer selector identifies the more probable correct one from \texttosql and \tableqa answers. 
Experiments show that both feature-based selector and LLM-based selector 
provide significant improvement over single models. 

\section{Table Question Answering Task}
Table Question Answering has received significant attention as it helps non-experts interact with complex tabular data.
%
%
Formally, given an input question $\mathcal{Q}=\{q_1,q_2,\dots,q_n\}$ and a table $\mathcal{T}$ with $\mathcal{R}$ rows and $\mathcal{C}$ columns, and each cell $\mathcal{T}_{i,j}$ contains a real value, \otableqa aims to produce an answer $\mathcal{A} = \{a_1,a_2,\dots,a_k\}$, where $q_n$ and $a_k$ are tokens. Then we introduce two main approaches for \otableqa: \texttosql and \tableqa.

\paragraph{\texttosql} Table QA problem is originally framed as semantic parsing, also known as \texttosql parsers, where a parser takes both question and table header as input, and predicts a SQL query that is directly executable to get the answer. Early neural sequence-to-sequence parsers~\citep{guo-towards, wang-rat, rubin-smbop} encode question/schema with attention mechanism and uses SQL grammar to guide the decoding process. Recent approaches take advantages of pre-trained models, and they either fine-tune~\cite{Execution-Guided, scholak-picard} or prompt~\cite{dail_sql, din} large models for Text-to-SQL parsing.


\paragraph{\tableqa} Several issues limit applying \texttosql parsers into real scenarios: training SOTA parsers require large amounts of expensive SQL annotations; existing parsers largely ignore the value of table contents. With the help of model pre-trained on large scale table corpus, recent works focus on end-to-end \otableqa that ignores generating SQL queries as an intermediate step and directly predicts the final answer through either fine-tune~\cite{tapex, zhao-reastap, jiang-omnitab} or prompt large models~\cite{few}.




\section{Evaluating \texttosql and \tableqa}
\label{sec:eval}
In this section, we evaluate existing \texttosql and \tableqa models on two benchmark datasets: \wtq and \wikisql.

\subsection{Experimental Setup}
\paragraph{Dataset.} \wtq comprises 22,033 instances with a diverse array of intricate questions and tables. \squall \citep{squall} annotates 11,276 \wtq instances with pre-processed tables and SQL queries.\footnote{We train \texttosql models on \squall and test on \wtq, as $20\%$ of \wtq questions lack SQL annotations and cannot be answered by \texttosql.} Compared with classic datasets, e.g., \wikisql and Spider \citep{yu-spider}, designed for SQL prediction on well-maintained databases, \wtq contains complex tables and questions which are difficult to answer with SQL queries. As a large portion of Spider tables does not have table content, we use \wikisql to validate the generalizability of our findings, which contains 80,654 instances.




\paragraph{Model and Metric.}  We evaluate SOTA models that have publicly available source code or APIs: \texttosql includes T5 \citep{t5}, GPT \citep{OpenAI_GPT4_2023}, and DIN-SQL \citep{din}; \tableqa includes \tapex \citep{tapex}, \omnitab \cite{jiang-omnitab}, and GPT \citep{OpenAI_GPT4_2023}. As \texttosql models often generate invalid SQL queries \citep{lin-bridging, scholak-picard}, we devise a post-processing module to screen table content, rectify query misspellings, identify the closest string values, and resolve mismatches. For fine-tuned models, we choose the \texttt{large} version. For fair comparison, we report answer string exact match (EM) accuracy.

\begin{figure}[t]
    \centering
    \includegraphics[width=0.49\textwidth]{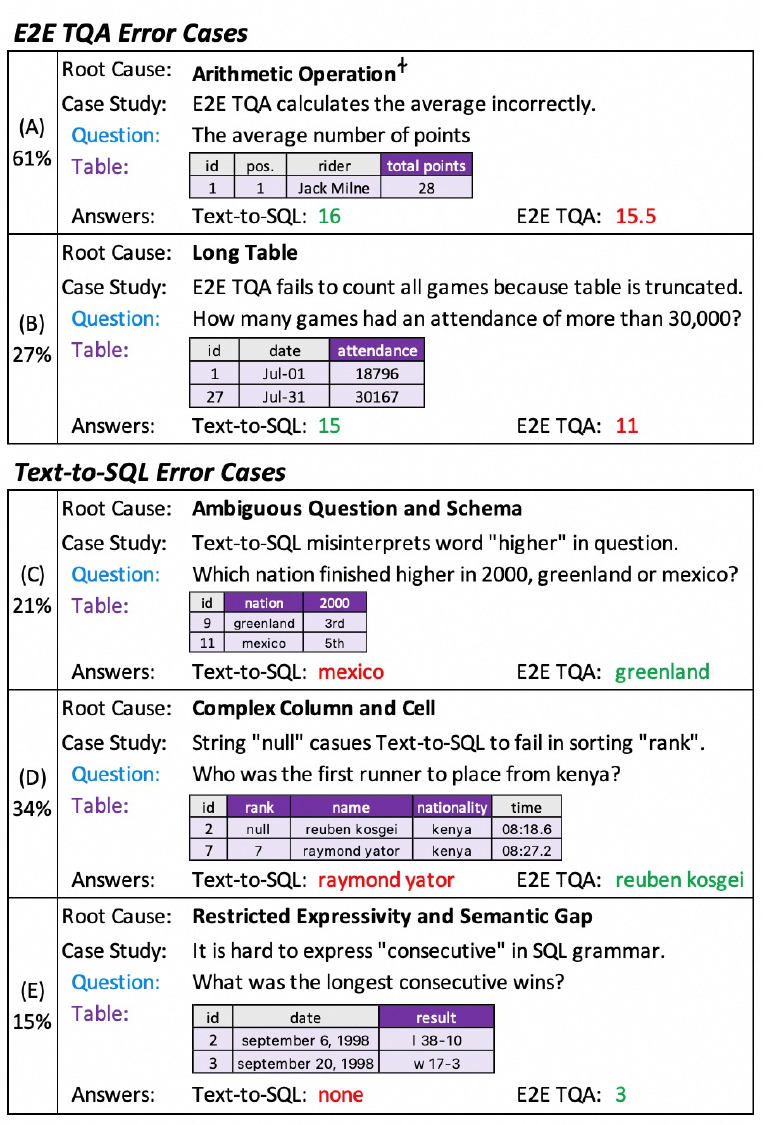}
\captionsetup{justification=justified,singlelinecheck=false}
    \caption{Error case analysis. $^\dagger$Arithmetic operation errors include questions with both long and short tables. Tables are regarded as long if their linearized sequences have more tokens than the \otableqa model input length. The percentage numbers on the left indicate the quantity of error cases, and remaining percentage points correspond to other errors, such incorrect labels.}
    \label{fig:errors}
\end{figure}

\subsection{Results}
\label{two_model_analysis}

According to Table \ref{experiments}, prompting methods (i.e., GPT and DIN-SQL) underperform fine-tuned models in table understanding on \wtq and \wikisql, aligning with findings in \citep{tablegpt, mixsc}. Thus, we primarily focus on fine-tuned \texttosql and \tableqa models. Best \texttosql and \tableqa models achieve comparable accuracy, but notably, 27.6\% of WTQ and 11.7\% of \wikisql questions were correctly answered exclusively by either \texttosql or \tableqa. It implies that models excel in tackling different types of table-based questions. To further investigate the strengths and weaknesses, we analyze 200 erroneous cases summarized in Figure \ref{fig:errors} (see detailed breakdown in Appendix \ref{pie}).

\paragraph{\texttosql is skilled at arithmetic operations.} It is evident in Figure \ref{fig:errors} (A) that 61\% of \tableqa error cases involve arithmetic operations including counting, summation, averaging, and subtraction. Despite existing \tableqa approaches~\citep{herzig-tapas, eisenschlos-understanding} have incorporated a separate aggregation operator into model design, the range of supported operations is limited with suboptimal performance. In contrast, \texttosql provides more accurate and consistent results for arithmetic operations through symbolic reasoning \citep{Binder, mixsc}.

\begin{figure}[t]
    \centering
\includegraphics[width=0.5\textwidth]{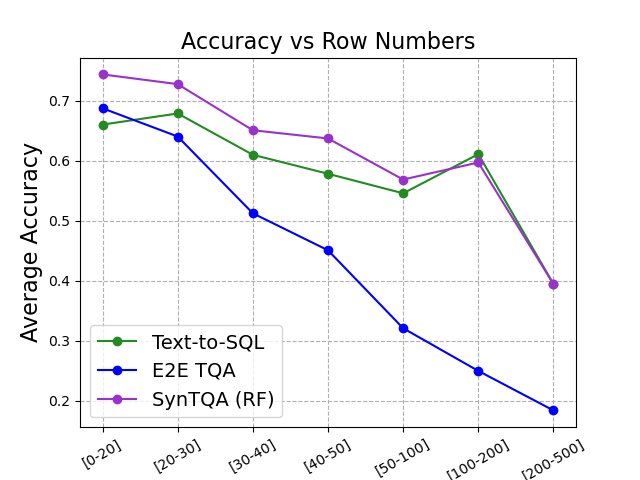}  \captionsetup{justification=justified,singlelinecheck=false}
    \caption{The impact of table size (i.e., number of rows) on the accuracy of \tableqa, \texttosql, and \syntableqa(RF) on the the test set of WTQ. The x-axis represents the row number ranges, and the y-axis shows the average accuracy for each method.}
    \label{fig:rows}
\end{figure}

\paragraph{\texttosql is adept at long tables.} When faced with long tables, comparing with \texttosql, \tableqa accuracy dramatically declines with increased table size (see details in Figure \ref{fig:rows} and Appendix \ref{rows}). This is because existing \tableqa approaches are limited in processing and understanding long context, therefore are only able to take truncated table as input. In contrast, \texttosql approaches primarily focus on table headers, and are more robust to incomplete or long table content. For example, in a long table like Figure \ref{fig:errors} (B), \texttosql is able to aggregate all critical information over the rows.

\paragraph{\tableqa is robust to ambiguous questions and non-standard table schema.} Rather than centering on table schema, \tableqa prioritizes table content. Analysing table content is particularly useful for resolving the ambiguity. As shown in Figure \ref{fig:errors} (C), the term ``\textmd{higher}'' may refer to a bigger value in quantity or a smaller value in rank. And \tableqa is more effective to infer that ``\textmd{higher}'' corresponds to a smaller ranking value by incorporating the table content (e.g., ``\textmd{3rd}'' and ``\textmd{5th}''). Instead, \texttosql relies on the relevant column header ``\texttt{2000}'' and mistakenly searches for the bigger value. Furthermore, as depicted in the third question of Figure \ref{fig:flow}, the non-standard header \texttt{tour} misleads \texttosql to retrieve the identity number. In contrast, \tableqa accurately predicts the official title of the tour.


\paragraph{\tableqa is flexible to process complex table content.} Complex content arises with the mixing of data types within same column, and \texttosql cannot find such nuanced difference, without looking at table contents. According to Figure \ref{fig:errors} (D), \texttosql cannot exclude the row with ``\textmd{null}'' rank and makes the wrong prediction then.

\paragraph{Some questions cannot map to a SQL query.}
As pointed out by \citep{squall}, there are cases where SQL queries are insufficiently expressive. 
According to Figure \ref{fig:errors} (E), \texttosql cannot answer questions related to phrases ``\textmd{approximate}'', while \tableqa is about to find the answers.

\paragraph{\texttosql requires post-process the executed answers.}
We also find that in some cases, additional steps are needed to translate SQL query results to natural language answers, where a notable semantic gap exists. For example, mapping ``\textmd{1}” to ``\textmd{longer}” for ``\textmd{longer or shorter}” question. \tableqa approaches do not have these limitations as they directly predict the final answers.

\section{\syntableqa: Selecting Correct Answer}
\label{sec:SynTableQA}



Above findings show that different models solve different questions, so we use a selector to choose the answer. Specifically, at each time, the selector receives the input of table $\mathcal{T}$, question $\mathcal{Q}$, \texttosql prediction and confidence $\widehat{\mathcal{A}}_{SQL}$, along with \tableqa prediction and confidence $\widehat{\mathcal{A}}_{E2E}$. Afterwards, the selector determines the correct answer $\widehat{\mathcal{A}}_{SEL}$, where $\widehat{\mathcal{A}}_{SEL} \in \{\widehat{\mathcal{A}}_{SQL}, \widehat{\mathcal{A}}_{E2E}\}$. In general, answer selection can be done through feature-based classification or LLM-based contextual reasoning, which is discussed in this section.

We use the best performing base models, i.e., fine-tuned T5 for \texttosql and \omnitab for \tableqa in the ensemble model.

\subsection{Selector Designs}

\paragraph{Feature-based Selector} \syntableqa(RF) trains a random forest classifier to make the selection.\footnote{We evaluate various classic classifiers and identify random forest as the top performer in Appendix \ref{classifiers}.} We design the following features to train the classifier: \textit{question characteristics} (e.g., question word and length), \textit{table characteristics} (e.g., table size, header and question overlapping, and truncation), \textit{\texttosql answer characteristics} (e.g., confidence, query execution, and queried answer data type), and \textit{\tableqa answer characteristics} (e.g., confidence and length). The full list of features and training details are included in Appendix \ref{features}. 

\paragraph{LLM-based Selector} \syntableqa(GPT) does not require training data thanks to LLMs' remarkable few-shot capabilities. For comparison, we evaluate LLMs' answer selection capability via direct prompting in Table \ref{experiments}.\footnote{We employ \texttt{gpt-3.5-turbo-0125} for the evaluation.} Furthermore, we propose a heuristic-enhanced prompting strategy to elevate the SOTA performance to 74.4\% and 93.6\% on \wtq and \wikisql (see details in Appendix \ref{llm-based selector}).



\begin{table}[t]
\renewcommand{\arraystretch}{1.1} 
\begin{tabular}{p{0.2\textwidth}cccc}
\toprule[1.5pt]
\textbf{Model}  & \multicolumn{2}{c}{\textbf{\wtq}}  & \multicolumn{2}{c}{\textbf{\wikisql}} \\ \cline{2-5}
  & Dev  & Test  & Dev & Test \\ \hline
\multicolumn{5}{c}{\textit{\texttosql Models}} \\
DIN-SQL   &  &  44.6 &   & 81.7 \\
GPT + TC + P   &   & 50.0  &  & 82.2 \\
T5 + TC + P   & 66.7  & \textbf{64.7}  & 88.3 & \textbf{89.6} \\ \hline
\multicolumn{5}{c}{\textit{\tableqa Models}}     \\
GPT      &   & 56.8  &  & 62.6 \\
\tapex               & 57.5 & 57.0 & 89.2 & \textbf{89.5} \\ 
\omnitab  & 63.7 & \textbf{62.6} & 89.7 & 89.0\\ 

\hline
\multicolumn{5}{c}{\textit{Ensemble Models}}    \\
\makecell[l]{\syntableqa(RF)}  &  & \textbf{71.6} & & \textbf{93.2}\\
\syntableqa(GPT)   & &  70.4  & &  93.0
\\
\syntableqa (Oracle)              &      & 77.5 &      & 95.1 \\ 

\bottomrule[1.5pt]
\end{tabular}
\captionsetup{justification=justified,singlelinecheck=false}
\caption{Accuracy on WTQ and \wikisql datasets comparing \syntableqa with baselines. The best test result is highlighted in \textbf{bold}. Oracle result indicates the maximum potential of mixing \texttosql and \tableqa models (TC: Table Content, P: Post-processing).}
\label{experiments}
\end{table}


\subsection{Results}



According to Table \ref{experiments} (Bottom), our ensemble models exhibit substantial improvement over individual models. They achieve comparable performance with recent tool-based LLMs on \wtq while saving computational costs, e.g., Dater \citep{dater} 65.9\% and Mix SC \citep{mixsc} 73.6\%. As our findings are orthogonal to these methods, we demonstrate a case integrating the concept of Mix SC in Appendix \ref{sc}. The effectiveness can be attributed to selectors' high success rate (nearly 80\%) in selecting correct answers.
Notably, the confidence of \texttosql and \tableqa models is the most impactful feature for \syntableqa(RF). 

\subsection{SQL Annotation Efficiency}
Since manually creating SQL annotations can be costly \citep{squall}, we conducted experiments to study how the accuracy improvement varies with different amounts of SQL annotations, using the feature-based selector in the WTQ dataset. The answers are assumed to be always fully available, leading to a stable performance of \tableqa. 

\begin{figure}[ht]
    \centering
    \includegraphics[width=0.45\textwidth]{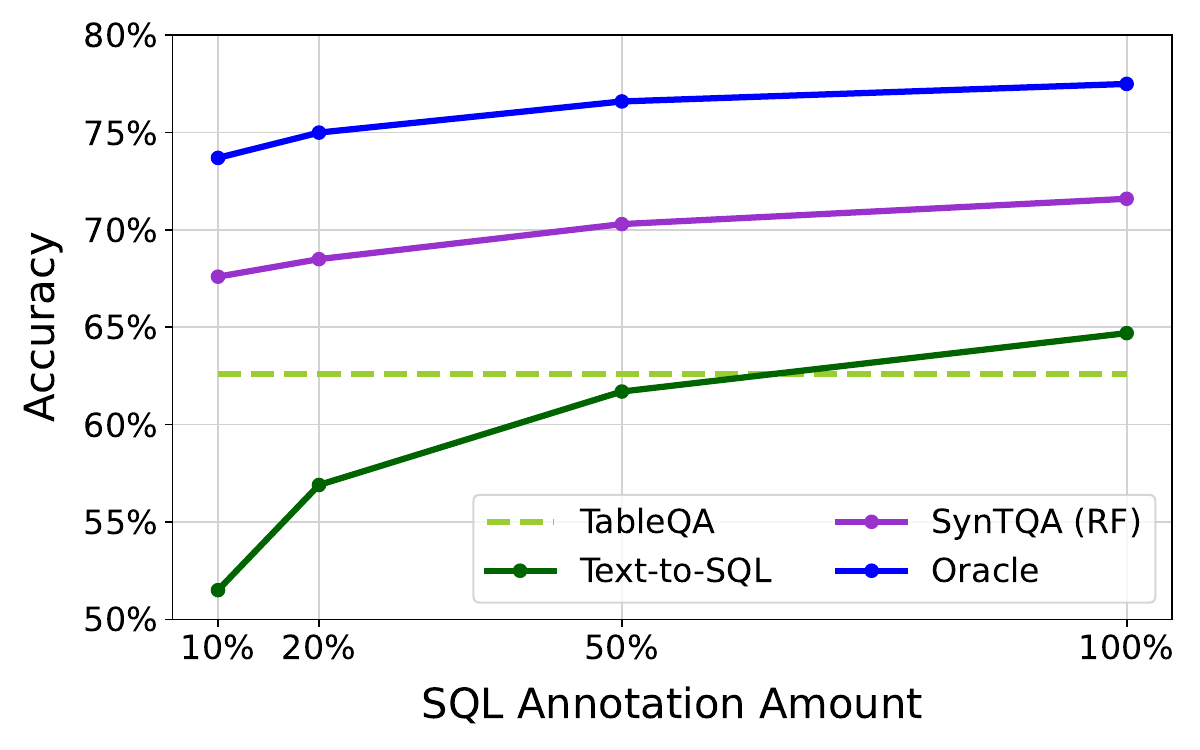}   \captionsetup{justification=justified,singlelinecheck=false}
    \caption{\wikisql test set accuracy versus the percentage amount of SQL annotations provided by \squall. Even an inferior \texttosql model trained with a more limited set of SQL annotations can substantially enhance the \tableqa model.}
    \label{fig:anno_eff}
\end{figure}

As shown in Figure \ref{fig:anno_eff}, 10\% of SQL annotations ($\sim$900) enhanced \tableqa accuracy by 5\%. The improvement potential and actual improvement continue to grow with the  increase of the SQL annotation amount. Trade-offs can be made between the performance improvement and annotation amounts depending on the use case.

\subsection{Robustness Analysis}

In addition to individual \texttosql and \tableqa models such as previous works \citep{pi, singha2023tabular}, we evaluate our ensemble approach \syntableqa (RF) with adversarial perturbations such as replacing key question entities and adding table columns. The evaluation is performed on the \robut-\wikisql dataset \citep{zhao-robut}. We find that different models exhibited degradation on distinct adversarial samples. Employing model assembling mitigates the performance degradation experienced by individual models significantly (see details in Table \ref{table:robut}).

\section{Other Related Work}

\paragraph{Mixture-of-Experts} Since proposed by  \citep{moe}, Mixture-of-Experts has been applied in a wide fields of machine learning \citep{li2022branchtrainmerge, gururangan2023scaling}. 
We follow the same concept and route the experts from the sample level \citep{Puerto2021MetaQACE, si-getting}, i.e. selecting an expert model for each test instance.

\paragraph{Tool-based LLMs} With LLMs' strong textual reasoning and tool-use capabilities, recent \otableqa methods \citep{Binder, dater, mixsc} call executable programs (e.g., SQL and Python) as needed to retrieve relevant contexts, facilitating reasoning. We provide an alternative ensemble approach that does not rely on computationally expensive LLMs. 




\section{Conclusion}

This study delved into the comparative analysis of two \otableqa approaches: \texttosql and \tableqa. Results indicate \texttosql's proficiency in arithmetic operations and long tables and \tableqa's advantages in resolving ambiguity and complexity in the question and table. We enhance performance on \otableqa datasets by combining models through answer selectors.
We plan to extend the method to more challenging problems such as hybrid TQA \citep{chen-hybridqa, zhu-tat}.



\section*{Limitations}
Although \omnitab is pre-trained for \tableqa, T5 is not a model specifically designed for \texttosql. Most \texttosql models are tailored for the Spider dataset \citep{Execution-Guided, rubin-smbop, scholak-picard}. Table or passage retrievers \citep{karpukhin-dense, dtr} can be applied to select certain rows and columns before truncating the long tables which might improve \tableqa performance. As for \syntableqa(GPT), we constrain GPT to select an answer from candidates, which abandons its capability to provide a different answer when both candidates are wrong. In more challenging datasets which necessitate both textual and tabular data \citep{chen-hybridqa, zhu-tat}, our method may not be as flexible and effective as tool-based LLMs \citep{li2023chain, asai2023self}.



\section*{Ethics Statement}
\syntableqa were developed using WTQ \citep{pasupat-compositional}, \squall \citep{squall}, \wikisql \citep{wikisql}, and \robut \cite{zhao-robut}, which are publicly available under the licenses of CC BY-SA 4.0\footnote{\url{https://creativecommons.org/licenses/by-sa/4.0/}}, BSD 3-Clause\footnote{\url{https://opensource.org/license/bsd-3-clause}}, and MIT\footnote{\url{https://opensource.org/license/mit}}. We used 4 NVIDIA Quadro RTX8000 GPUs to fine-tune models. \syntableqa(RF) and \syntableqa(GPT) were constructed and executed solely using CPU. \syntableqa(GPT) relies on OpenAI API and using other GPT versions will lead to varied performance. No manual annotation and human study are involved in this study.

\section*{Acknowledgements}
This research is supported by the RIE2025 Industry Alignment Fund – Industry Collaboration Projects (IAF-ICP) (Award I2301E0026), administered by A*STAR, as well as supported by Alibaba Group and NTU Singapore. Siyue Zhang and Chen Zhao were supported by Shanghai Frontiers Science Center of Artificial Intelligence and Deep Learning, NYU Shanghai.
This work was supported in part through the NYU IT High Performance Computing resources, services, and staff expertise.



\clearpage
\appendix

\section{Evaluation Implementation Details}
\label{implement}

To fully utilize the table-question-answer triplets from WTQ and SQL annotations from \squall, we augmented the random splits generated by \squall with additional WTQ samples that were not annotated within \squall. In the evaluation, we used the split of \texttt{train-1} for fine-tuning \texttosql and the corresponding augmented split to fine-tune \tableqa. Then, both fine-tuned models are evaluated by the augmented \texttt{dev-1} set. Specifically, the training set comprises 11,340 WTQ samples, with SQL annotations present in 9,032 of them. As for \wikisql, we employed the full dataset with 56,640 table-question-SQL query training samples. Answers were extracted following the approach outlined in \cite{tapex}. We used the default split for the evaluation, named as \texttt{train-0} and \texttt{dev-0}. For model fine-tuning, we maintained the same parameters as original papers, running 50 and 10 epochs for WTQ and \wikisql and selecting the best checkpoint based on the validation accuracy.

\section{Statistics of Error Cases}

We analyse 200 error cases for \texttosql and \tableqa models. The detailed breakdown is shown in Figure \ref{pie}. The remaining percentage points correspond to other errors, such incorrect labels.

\label{pie}
\begin{figure}[H]
    \centering
\includegraphics[width=0.27\textwidth]{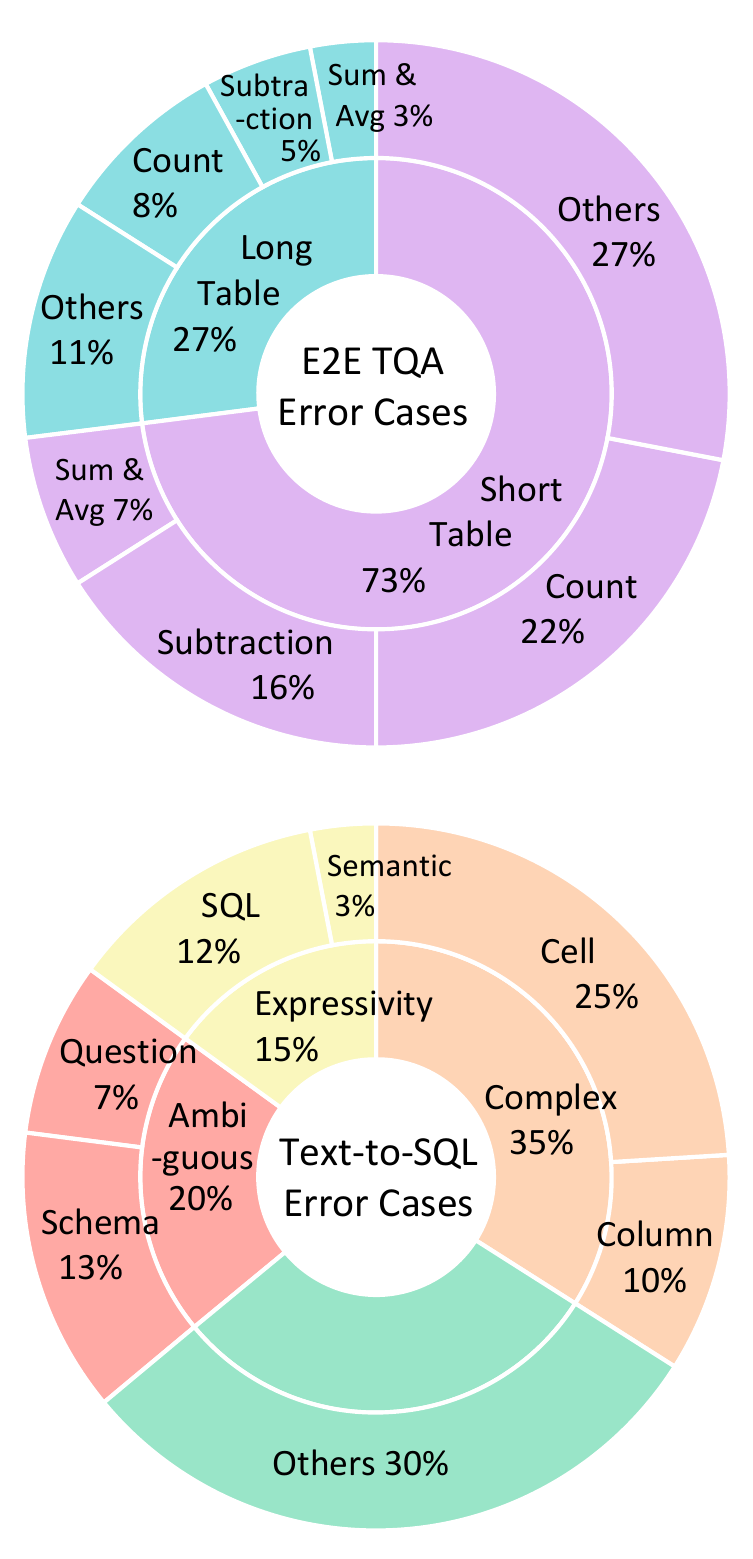}  \captionsetup{justification=justified,singlelinecheck=false}
    \caption{Breakdown of \tableqa error cases (top) and  \texttosql error cases (bottom).}
    \label{fig:pie}
\end{figure}

\section{Table Size Impact Analysis}
\label{rows}

This section analyzes how table size, measured by row numbers, influences the performance of various methods on WTQ. We investigate the impact of row count on the average accuracy of \tableqa, \texttosql, and an ensemble approach. Our findings reveal a consistent trend of decreasing accuracy as the number of rows increases. Notably, \tableqa experiences a more pronounced decline in accuracy compared to \texttosql. Traditional \texttosql methods typically rely solely on table schema for SQL generation, leading to consistent accuracy regardless of the number of rows. However, the decline of \texttosql accuracy shown in Figure \ref{fig:rows} implies that table content may also play a role in SQL generation. Besides, it is evident that \tableqa deteriorates much more severely than \texttosql which can be attributed to the lack of the retrieval system (i.e., table row and column selection) and the the complexity of handling long-context data. Last but not least, the ensemble approach is observed to be effective to mitigate the accuracy drop caused by the table size.




\section{LLM-based \otableqa Models}
\label{llm_eval}
This section presents the evaluation of LLM-based \tableqa and \texttosql models. To optimize the cost, we use \texttt{gpt-3.5-turbo-0125} for all models including \tableqa, \texttosql, and \syntableqa(GPT) selector. For LLM-based \tableqa, we follow the direct prompting (zero-shot) approach implemented by \citep{mixsc}. For LLM-based \texttosql, we incorporate 8 examples from \texttt{dev} set in the prompting to showcase the target style of SQL queries. 


Table \ref{gpt-experiments} demonstrates that GPT-3.5 exhibits limited proficiency in table understanding, as evidenced by significantly lower accuracy of both GPT-based \texttosql and \tableqa models compared to fine-tuned small models. However, it is evident that GPT-based \texttosql and \tableqa models also response correctly to different questions, mirroring the findings observed between T5 and \omnitab. The gap between the oracle accuracy and individual model accuracy suggests the substantial improvement potential by aggregation.

\begin{table}[ht]
\renewcommand{\arraystretch}{1.1} 
\begin{tabular}{p{0.25\textwidth}cc}
\toprule[1.5pt]
\textbf{Model} & \textbf{WTQ}  & \textbf{WikiSQL} \\ 
\hline
\multicolumn{1}{c}{\textit{\texttosql Models}} \\
T5 & 67.6  & 90.8 \\ 
GPT & 50.0 & 82.2 \\
\hline
\multicolumn{1}{c}{\textit{\tableqa Models}}     \\
\omnitab  & 66.3 & 88.3 \\  
GPT & 56.8 & 62.6 \\
\hline
\multicolumn{1}{c}{\textit{Ensemble Models}}    \\
\syntableqa(GPT)   &  65.2  & 84.4 \\
\syntableqa (Oracle)    &  75.2  & 87.6 \\ 
\bottomrule[1.5pt]
\end{tabular}
\captionsetup{justification=justified,singlelinecheck=false}
\caption{Accuracy on subsets of WTQ and \wikisql. \syntableqa aggregates LLM-based \texttosql and \tableqa models via the LLM-based selector. Oracle result indicates the maximum potential of mixing LLM-based \texttosql and \tableqa models.}
\label{gpt-experiments}
\end{table}

\newpage
\section{Feature-based Selector Implementation}
\label{features}

\subsection{Classifier Features}
Below we list all the features used to train our random forest classifier for selecting the final output answer based on model predictions.

\begin{itemize}
    \item \textbf{Question Characteristics}: question word, question length, and the number of numerical values in the question. 
    \item \textbf{Table Characteristics}: the number of rows and columns in the table, the number of overlap words between the table header and question, and a boolean value implying whether the table is truncated in the model input. 
    \item \textbf{\texttosql Answer Characteristics}: with regard to the predicted and revised SQL query, it includes the generation probability normalized by length, and the number of preprocessed columns used in the query (e.g., \texttt{\_parsed}, \texttt{\_first}, and \texttt{\_list} in \squall); concerning the queried answers from the table, it consists of the query execution status (i.e., successful or not), the number of queried answers, and the data types of queried answers (i.e., string or number).
    \item \textbf{\tableqa Answer Characteristics}: the generation probability normalized by length, the number of predicted answers, answer data types, a boolean value indicating whether the \tableqa answer is a sub-string of the \texttosql answer, and another boolean indicator checking if the \tableqa answer is a sub-string of the model input.
\end{itemize}

\subsection{Training Details}

Error case samples, where one model is correct and the other one is erroneous, are essential for effectively training the random forest classifier. Thus, we trained one \texttosql model and one \tableqa model at a time for each dataset splitting (in total 5 splits). We gathered error cases from each validation set. As \wikisql does not provide 5 random splits as \squall, 4 additional unique \texttt{dev} sets with a similar amount of samples as the original \texttt{dev} set were extracted from the \texttt{train} set.

\subsection{Comparisons Among Classifiers}
\label{classifiers}

We investigate various classic classification methods for answer selection in \syntableqa: linear regression (LR), k-nearest neighbors (kNN), support vector machine (SVM), multilayer perceptron (MLP), and random forest (RF). As shown in Table \ref{cls}, RF attains the best performance in answer selection.

\begin{table}[h]
\begin{tabular}{p{0.1\textwidth}cccccc}
\toprule[1.5pt]
Model    & LR   & kNN  & SVM   & MLP  & RF            \\ \hline
Accuracy & 70.8 & 66.8 & 70.1  & 70.0 & \textbf{71.6} \\ 
\bottomrule[1.5pt] 
\end{tabular}
\captionsetup{justification=justified,singlelinecheck=false}
\caption{Classification accuracy of different machine learning methods in \syntableqa on the test set of \wtq dataset. The best performance is highlighted in \textbf{bold}.}
\label{cls}
\end{table}

\section{Heuristic Enhanced \syntableqa(GPT)}
\label{llm-based selector}

\begin{figure}[h]
    \centering
    \includegraphics[width=0.45\textwidth]{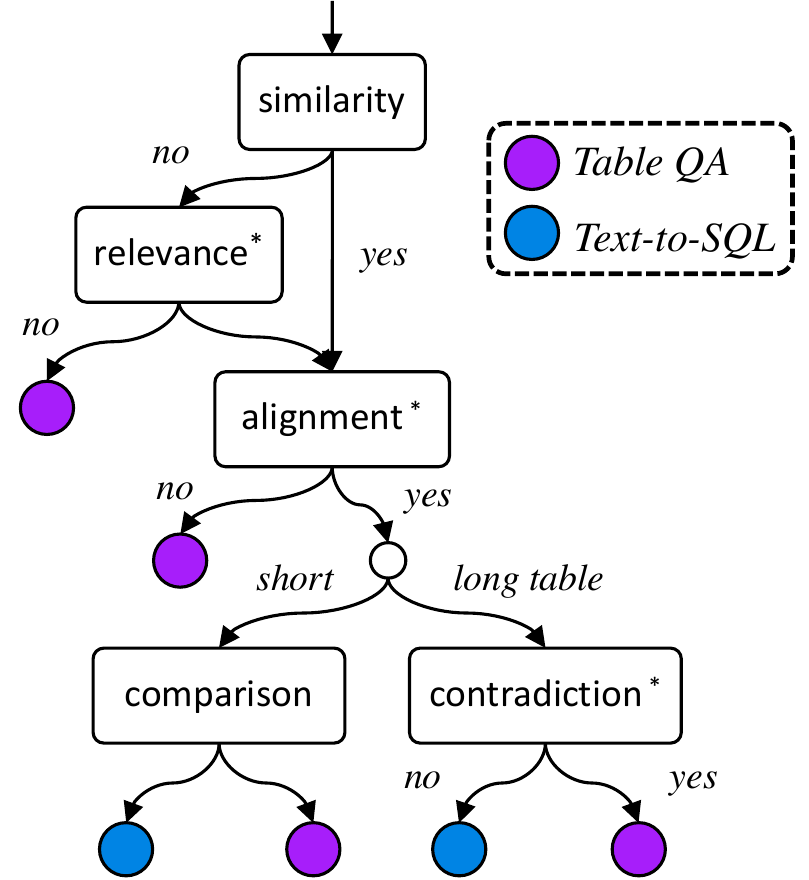}   \captionsetup{justification=justified,singlelinecheck=false}
    \caption{Design of LLM-based Selector. \textit{Similarity} module examines if \texttosql and \tableqa answers are similar entities. \textit{Relevance} module checks if the \texttosql answer is relevant to the question. \textit{Alignment} module inspects if the number of entities in \texttosql answer corresponds to the question. \textit{Comparison} module chooses the correct answer from two models. \textit{Contradiction} module identifies if there is contradiction between the truncated table and \texttosql answer ($^*$indicates only using the \texttosql answer).}
    \label{fig:logic}
\end{figure}

Apart from the direct prompting approach presented in the paper, we also develop a heuristic-enhanced prompting strategy for \syntableqa(GPT) and test it with \texttt{gpt-4-0125-preview}. The main idea is to leverage additional LLM-based modules to reduce the the necessity of complex reasoning on the question, table, and answer candidates. The designed heuristic is demonstrated in Figure \ref{fig:logic} and the full prompts refer to the following subsections. As a result, the heuristic-enhanced prompting strategy achieves 89\% and 87.1\% accuracy in selecting the correct answer on \wtq and \wikisql respectively. Correspondingly, it attains \otableqa accuracy of 74.4\% and 93.6\% on \wtq and \wikisql, further elevating the SOTA \otableqa performance.

\subsection{Similarity Module}
\begin{figure}[H]
    \centering
    \includegraphics[width=0.48\textwidth]{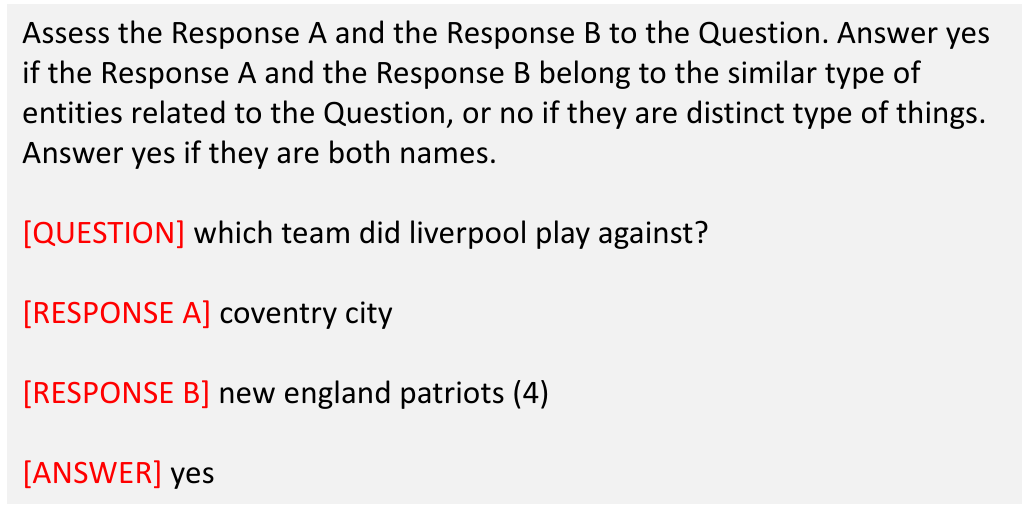}
\end{figure}

\subsection{Relevance Module}
\begin{figure}[H]
    \centering
    \includegraphics[width=0.48\textwidth]{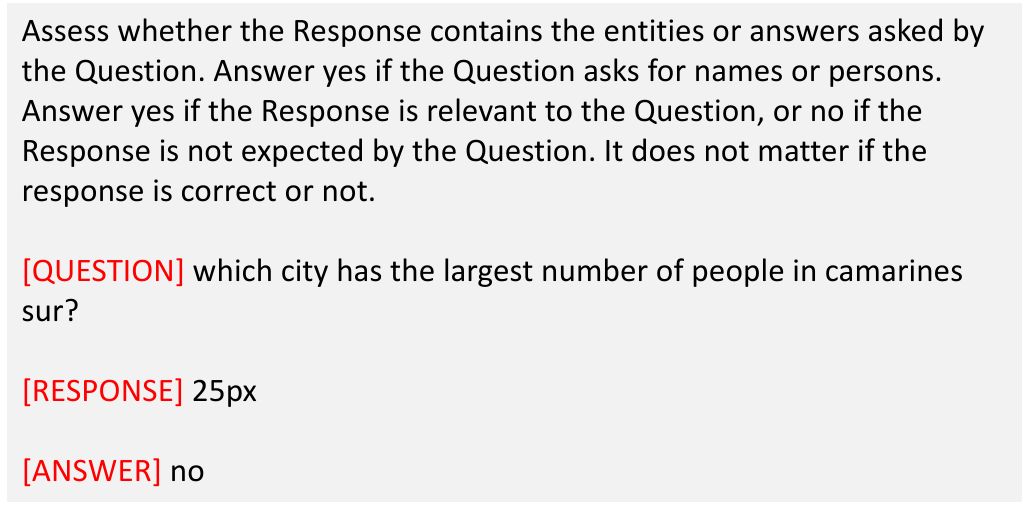}
\end{figure}

\subsection{Alignment Module}
\begin{figure}[H]
    \centering
    \includegraphics[width=0.48\textwidth]{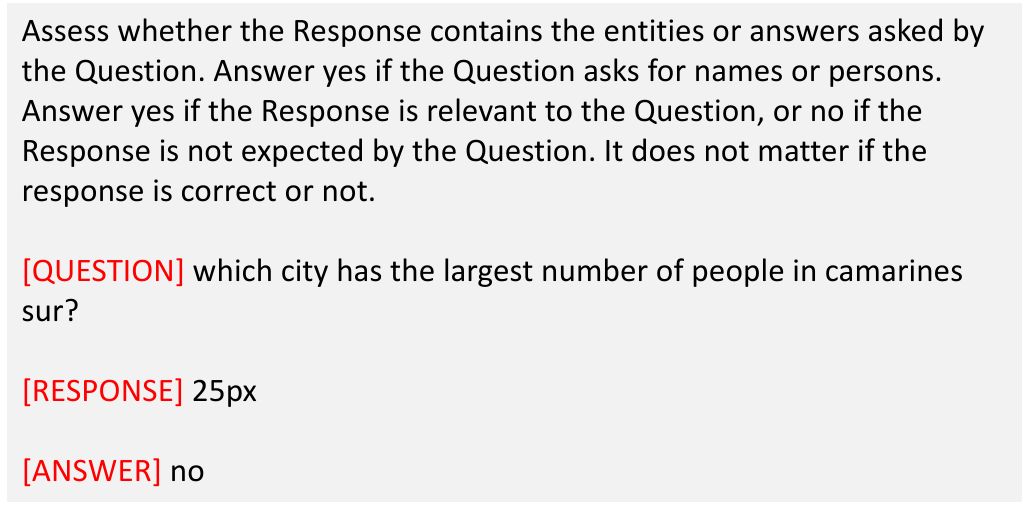}
\end{figure}

\subsection{Comparison Module}
\label{compare}

\begin{figure}[H]
    \centering
    \includegraphics[width=0.48\textwidth]{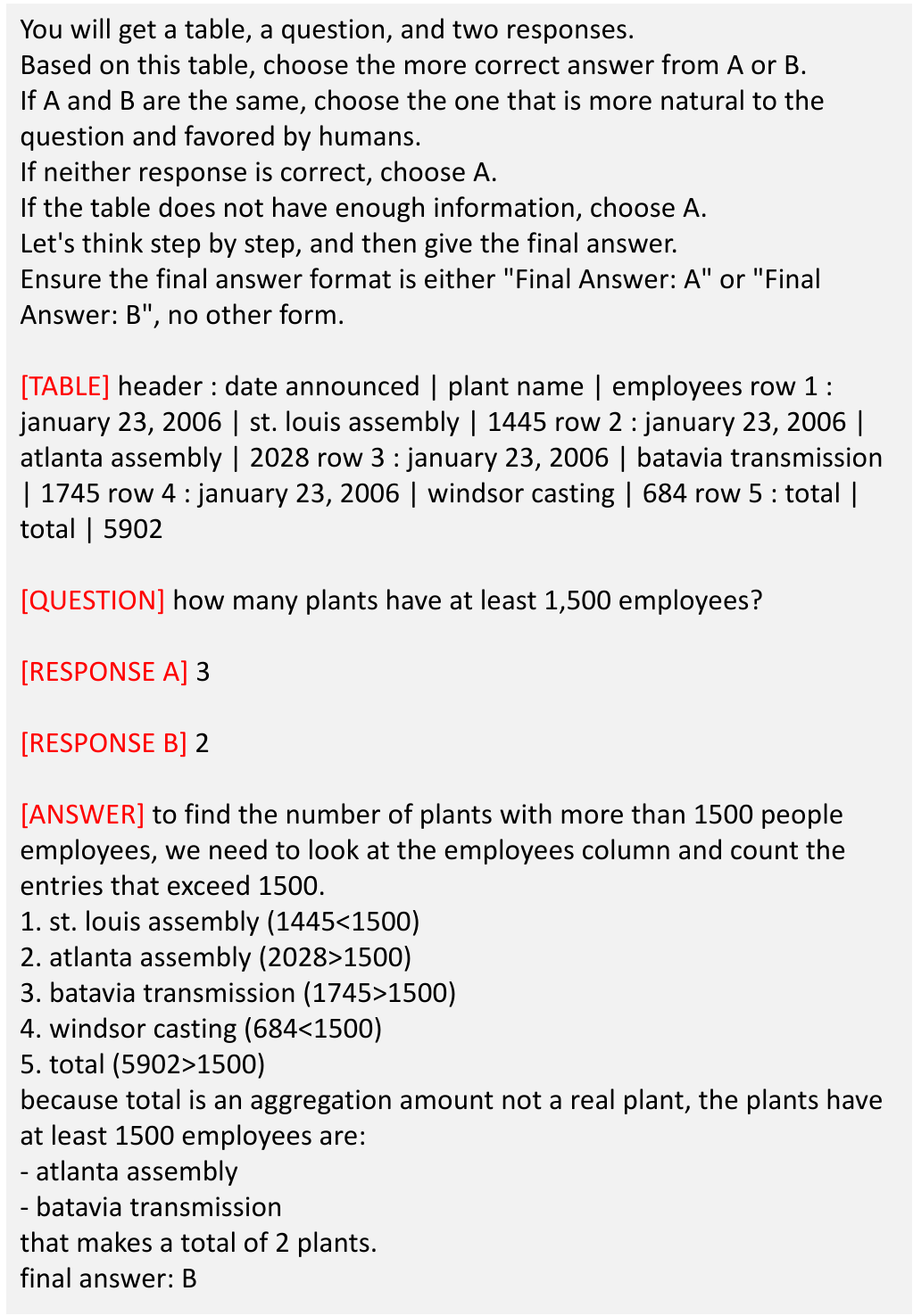}
\end{figure}

\subsection{Contradiction Module}
We implemented one type of contradiction scenarios regarding the question for counting entities in the table when candidate answers are small integer numbers. In the event that this module detects a higher count of entities within the truncated table than reflected in the \texttosql response, it is deemed a contradiction, indicating a high likelihood of errors within the response.

\begin{figure}[H]
    \centering
    \includegraphics[width=0.48\textwidth]{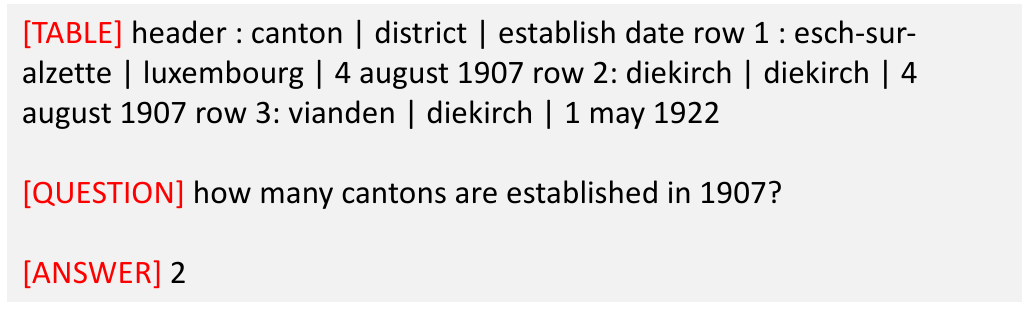}
\end{figure}

\section{Integrating Self-Consistency}
\label{sc}

Following \citep{mixsc}, we incorporate the Self-Consistency in our \texttosql and \tableqa base models. To generate 5 candidate answers for each model, we perturb the input table schema for the \texttosql model and conduct top-k sampling ($k=50$) for the \tableqa model. Among five candidates, we choose one following the rule of maximum voting. Lastly, our RF classifier determined the final answer based on designed features.

\begin{table}[h]
\renewcommand{\arraystretch}{1.1} 
\begin{tabular}{p{0.35\textwidth}c}
\toprule[1.5pt]
\textbf{Model}  & \textbf{\wtq}    \\ \hline
\multicolumn{1}{c}{\textit{\texttosql Models}} \\
T5     & 64.7   \\ 
T5 + SC   & 65.2   \\ \hline
\multicolumn{1}{c}{\textit{\tableqa Models}}     \\
\omnitab  &  62.6 \\ 
\omnitab + SC  & 62.9 \\ \hline
\multicolumn{1}{c}{\textit{Ensemble Models}}    \\
\makecell[l]{\syntableqa(RF)}   & 71.6 \\

\syntableqa (RF) + SC     & 71.8  \\ 

\bottomrule[1.5pt]
\end{tabular}
\captionsetup{justification=justified,singlelinecheck=false}
\caption{Accuracy on WTQ test set. Self-Consistency can further improve the performance of both individual models and the ensemble model (SC: Self-Consistency).}
\label{sc_experiments}
\end{table}

\onecolumn
\section{Robustness Analysis}
\label{robut}


\begin{table*}[h]
\renewcommand{\arraystretch}{1.1}
\begin{tabular}{lm{7.5em}cccccc}
\toprule[1.5pt]
&
&
\multicolumn{2}{c}{\texttosql} &
\multicolumn{2}{c}{\tableqa} &
\multicolumn{2}{c}{\syntableqa(RF)} \\ \cline{3-4} \cline{5-6} \cline{7-8}
\multirow{-2}{*}{Level} &
\multirow{-2}{*}{Perturbation Type} &
\textsc{Acc} &
\textsc{R-Acc} &
\textsc{Acc} &
\textsc{R-Acc} &
\textsc{Oracle} &
\textsc{Acc}  \\ \hline
&
Synonym \newline Replacement
&
{\color[HTML]{333333} \begin{tabular}[c]{@{}c@{}}84.7 / 72.6\\ \color{red}{(-12.1)}\end{tabular}} 
&
82.9
&
\begin{tabular}[c]{@{}c@{}}84.7 / 73.0\\ \color{red}{(-11.7)}\end{tabular}   
&
83.4
&
\begin{tabular}[c]{@{}c@{}}93.1 / 86.5\\ \color{red}{(-6.6)}\end{tabular}  

&
\begin{tabular}[c]{@{}c@{}}\textbf{79.6}\\ \color{Green}{(+6.6)}\end{tabular} 
   \\ \cline{2-8} 
\multirow{-3}{*}{\begin{tabular}[c]{@{}l@{}}Table\\ Header\end{tabular}}
&
Abbreviation \newline Replacement
&
\begin{tabular}[c]{@{}c@{}}84.4 / 76.2\\ \color{red}{(-8.2)}\end{tabular} &
87.0
&
\begin{tabular}[c]{@{}c@{}}84.2 / 74.3\\ \color{red}{(-9.9)}\end{tabular}
&
85.7 
&
\begin{tabular}[c]{@{}c@{}}92.9 / 87.5\\ \color{red}{(-5.4)}\end{tabular}
&
\begin{tabular}[c]{@{}c@{}}\textbf{81.2}\\ \color{Green}{(+5.0)}\end{tabular} 
\\ \hline
&
Column \newline Extension
&
\begin{tabular}[c]{@{}c@{}}89.6 / 48.7\\ \color{red}{(-40.9)}\end{tabular} 
&
52.9
&
\begin{tabular}[c]{@{}c@{}}91.6 / 54.8\\ \color{red}{(-36.8)}\end{tabular}
&
59.1
&
\begin{tabular}[c]{@{}c@{}}95.5 / 58.5\\ \color{red}{(-37.0)}\end{tabular}
&
\begin{tabular}[c]{@{}c@{}}\textbf{56.3}\\ \color{Green}{(+1.5)}\end{tabular} 
\\ \cline{2-8} 
\multirow{-3}{*}{\begin{tabular}[c]{@{}l@{}}Table\\ Content\end{tabular}} 
&
Column \newline Adding
&
\begin{tabular}[c]{@{}c@{}}81.0 / 79.7\\ \color{red}{(-1.3)}\end{tabular} &
94.7
&
\begin{tabular}[c]{@{}c@{}}81.5 / 70.3\\ \color{red}{(-11.2)}\end{tabular}
&
83.4
&
\begin{tabular}[c]{@{}c@{}}90.7 / 87.5\\ \color{red}{(-3.2)}\end{tabular}
&
\begin{tabular}[c]{@{}c@{}}\textbf{83.8}\\ \color{Green}{(+4.1)}\end{tabular} 
\\ \hline
&
Word-Level \newline Paraphrase
&
\begin{tabular}[c]{@{}c@{}} 87.3 / 63.7 \\ \color{red}{(-23.6)}\end{tabular} &
70.6
&
\begin{tabular}[c]{@{}c@{}}88.3 / 66.0\\ \color{red}{(-22.3)}\end{tabular}
&
72.9 
&
\begin{tabular}[c]{@{}c@{}}94.3 / 73.8\\ \color{red}{(-20.5)}\end{tabular}
&
\begin{tabular}[c]{@{}c@{}}\textbf{68.8}\\ \color{Green}{(+2.8)}\end{tabular} 
\\ \cline{2-8} 
\multirow{-3}{*}{Question} 
&
Sentence-Level Paraphrase
&
\begin{tabular}[c]{@{}c@{}}83.6 / 71.5\\ \color{red}{(-12.1)}\end{tabular} &
81.3
&
\begin{tabular}[c]{@{}c@{}}83.8 / 72.3\\ \color{red}{(-11.5)}\end{tabular}
&
83.1
&
\begin{tabular}[c]{@{}c@{}}92.2 / 83.7\\ \color{red}{(-8.5)}\end{tabular}
&
\begin{tabular}[c]{@{}c@{}}\textbf{78.0}\\ \color{Green}{(+5.7)}\end{tabular} 
\\ \hline
Mix &
--- &
\begin{tabular}[c]{@{}c@{}}87.0 / 60.3\\ \color{red}{(-26.7)}\end{tabular} 
&
66.8
&
\begin{tabular}[c]{@{}c@{}}88.5 / 63.4\\ \color{red}{(-25.1)}\end{tabular}
&
69.5
&
\begin{tabular}[c]{@{}c@{}}94.0 / 72.0\\ \color{red}{(-22.0)}\end{tabular}
&
\begin{tabular}[c]{@{}c@{}}\textbf{66.8}\\ \color{Green}{(+3.4)}\end{tabular} 
\\ 
\bottomrule[1.5pt]
\end{tabular}
\captionsetup{justification=justified,singlelinecheck=false}
\caption{Robustness evaluation results of \texttosql, \tableqa, and \syntableqa models on \robut-\wikisql. \textsc{Acc} represents the \emph{Pre-} and \emph{Post-perturbation Accuracy}; \textsc{R-Acc} represents the \emph{Robustness Accuracy}. \textbf{Bold} numbers indicate the highest \emph{Post-perturbation Accuracy} in each perturbation type. \textcolor{red}{Red} numbers show the accuracy degeneration due to the perturbation. \textcolor{Green}{Green} numbers demonstrate the improvement over the best individual model.}
\label{table:robut}
\end{table*}

\end{document}